\documentclass[10pt,twocolumn,letterpaper]{article}

\usepackage{cvpr}
\usepackage{times}
\usepackage{epsfig}
\usepackage{graphicx}
\usepackage{amsmath}
\usepackage{amssymb}

\usepackage{graphicx}
\usepackage{paralist}
\usepackage{booktabs}
\usepackage{multirow}
\usepackage{subfigure}

\newcommand{\fig}{Fig.~}
\newcommand{\eq}{Eq.\,}
\newcommand{\sect}{Section~}
\newcommand{\tab}{Table~}
\newcommand{\ap}{Appendix~}

\newcommand{\ba}{\boldsymbol{a}}
\newcommand{\bd}{\boldsymbol{d}}
\newcommand{\by}{\boldsymbol{y}}
\newcommand{\bp}{\boldsymbol{p}}
\newcommand{\bx}{\boldsymbol{x}}

\newcommand{\cardinality}[1]{{#1}}

\DeclareMathOperator*{\argmax}{arg\,max}
\DeclareMathOperator*{\softmax}{softmax}
\DeclareMathOperator*{\dist}{dist}

\newcommand{\ra}[1]{\renewcommand{\arraystretch}{#1}} 
\graphicspath{ {./figures/} }

\usepackage[breaklinks=true,bookmarks=false]{hyperref}

\cvprfinalcopy 

\def\cvprPaperID{****} 

\begin{document}

\title{Visual Question Answering with Prior Class Semantics}

\author{Violetta Shevchenko, Damien Teney, Anthony Dick, Anton van den Hengel \\
The University of Adelaide\\
{\tt\small \{violetta.shevchenko,damien.teney,anthony.dick,anton.vandenhengel\}@adelaide.edu.au}
}

\maketitle

\begin{abstract}
We present a novel mechanism to embed prior knowledge in a model for visual question answering.
The open-set nature of the task is at odds with the ubiquitous approach of training of a fixed classifier.
We show how to exploit additional information pertaining to the semantics of candidate answers.
We extend the answer prediction process with a regression objective in a semantic space, in which we project candidate answers using prior knowledge derived from word embeddings.
We perform an extensive study of learned representations with the GQA dataset, revealing that important semantic information is captured in the \emph{relations} between embeddings in the answer space.
Our method brings improvements in consistency and accuracy over a range of question types.
Experiments with novel answers, unseen during training, indicate the method's potential for open-set prediction.
\end{abstract}

\section{Introduction}

The task of visual question answering (VQA) has become a benchmark to evaluate joint progress in computer vision and natural language processing. This complex task, in its most general formulation, requires deep analysis of both visual and textual information in order to correctly answer a question, given an associated image. Behind its simple formulation, VQA is an extremely complex task that offers a testbed for a multitude of capabilities required to develop strong AI systems.

Most recent developments in the field of VQA have focused on the development of deep learning architectures that can be trained with end-to-end supervision (\ie questions, images, and answers). However, even current large-scale datasets
~\cite{antol2015vqa,goyal2017making} can only cover a limited fraction of all knowledge potentially useful for the task. The underlying reasons for this limitation are that
\begin{inparaenum}[1)]
  \item the collection of data with end-to-end annotations, \ie questions/answers is expensive as it usually requires human resources,
  \item the desirable knowledge about the world is constantly expanding, and no single dataset can ever capture it all.
\end{inparaenum}
Existing models trained once and for all on any of these datasets lack the generalization and adaptation capabilities desirable in real-world applications. These shortcomings motivate our search for alternative sources of information, and a method to exploit them in a VQA model.

\begin{figure}[t]
  \centering
  \includegraphics[width=1\linewidth]{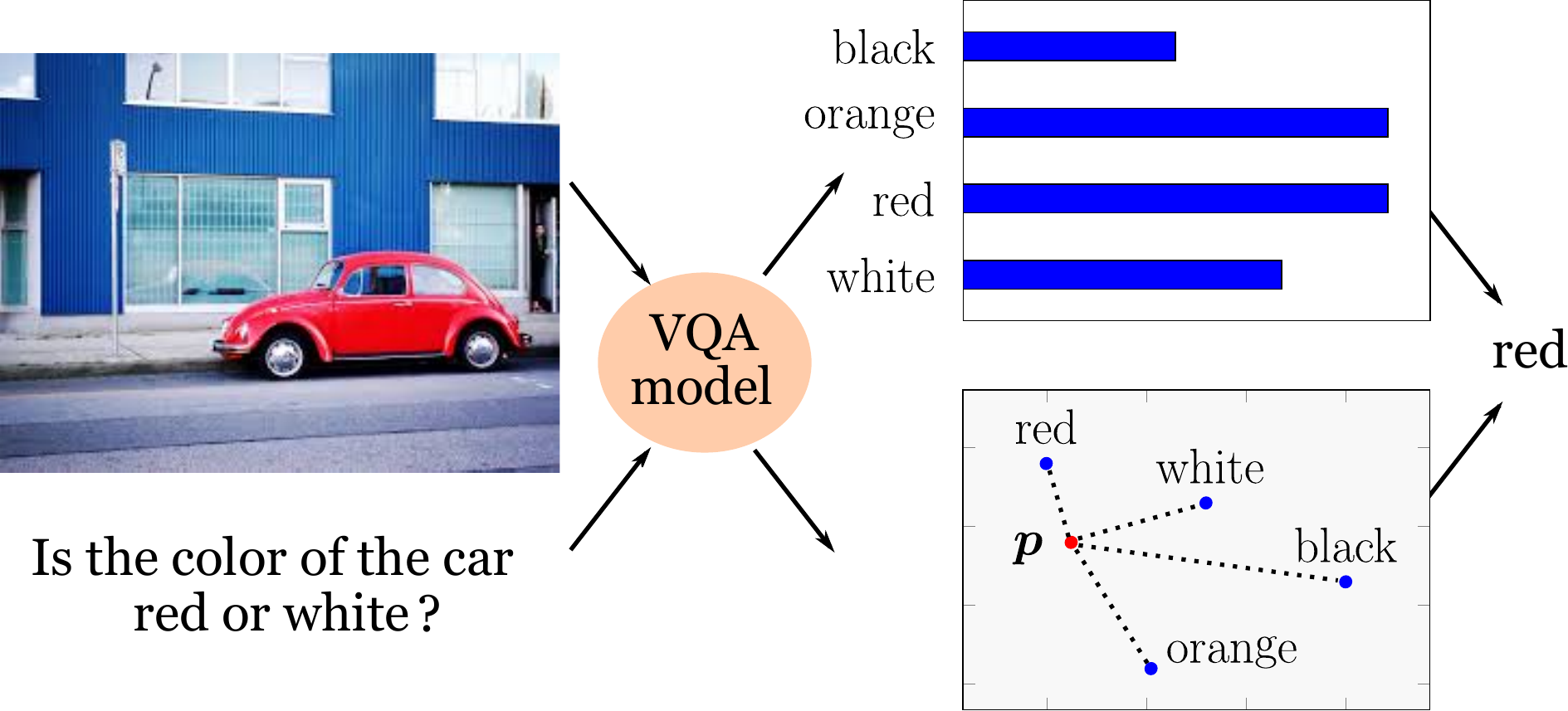}
  \caption{Existing models treat VQA as a classification task over predefined answers (upper branch). We supplement our model with a regression objective in a semantic answer space (lower branch). This allows incorporating additional prior knowledge about answer semantics. This improves its accuracy and consistency. In the above example, \textit{red} and \textit{orange} are similarly likely with the traditional objective. Our regression lands closer to the representation of \textit{red} in the answer space. This resolves the ambiguity and \textit{red} is chosen as the final answer.} 
  \label{fig:scheme}
  \vspace{-11pt}
\end{figure}

A common approach to include existing knowledge in VQA models is to use pretrained models to obtain image and question features. On the image side, pretrained convolutional neural networks (CNNs) or object detectors are ubiquitous~\cite{anderson2018bottom} to extract representative image features. On the language side, pretrained word embeddings like Word2Vec~\cite{mikolov2013efficient} and GloVe~\cite{pennington2014glove} usually serve to encode the words of the question. The advantage of these techniques is to leverage knowledge learned from larger, non-VQA specific data (\eg ImageNet and large text corpora). The benefit of these approaches has been widely demonstrated, which further motivates our quest for additional sources of usable knowledge and techniques to incorporate it.

Existing models for VQA follow the common blueprint of a two-stream embedding, followed by fusion and classification stages~\cite{antol2015vqa,teney2018tips,yang2016stacked}. The typical setting in VQA consists of an image and a related question. The model takes this image-question pair and predicts the correct answer by solving a classification problem over the set of candidate answers that occur in the training data. This classification approach, in contrast to text generation~\cite{wu2017image,gao2015you}, considerably simplifies the evaluation process, as the model can be assessed by its classification accuracy. However, treating VQA as a classification task has major drawbacks. The answers are treated as distinct class labels and answer words are abstracted from their meanings. This disregards semantic relations between related answers. Moreover, some questions contain possible answers in their wording (\eg~\textit{Is this car red or white~?}) and it seems natural to include mechanisms to explicitly represent the semantics of possible answers as done for question words. Guided by these observations, we develop an architecture that leverages prior knowledge about answer to improve the performance of a VQA model.

Our main technical contribution is to treat VQA as a multitask problem, where we both predict the answer label based on classification scores, and we additionally learn a mapping into an answer representation space that captures the semantics of these answers (see \fig\ref{fig:scheme}). We incorporate prior knowledge into the model by initializing the representations of answers with pretrained word embeddings. We perform an extensive and rigorous analysis of the trained model. It demonstrates the benefits of the approach and provides us with insights in the ways language semantics are useful for the task of VQA. Moreover, we show that learned answer representations can be used for out-of-vocabulary answer prediction which is an important, yet understudied problem in VQA field~\cite{noh2019transfer}. 

\vspace{4pt}
The contributions of this paper are as follows.
\vspace{-\topsep}
\vspace{-1pt}
\begin{itemize}
  \setlength\itemsep{-1.5pt}
  \item We formulate VQA as a multitask problem, where we train the model, not only to assign scores to answer candidates, but also to perform a regression in a vector space that represents answer semantics.
  \item We use this multitask formulation to incorporate additional information into the model with a particular loss and initialization of the semantic answer space. We also show that it allows the model to predict novel answers that were not seen during training.
  \item We perform an extensive analysis of the model and various ablations. We demonstrate clear advantages on the GQA dataset~\cite{hudson2018gqa}, and obtain insights on the ways in which answer semantics are useful for the task of VQA.
\end{itemize}
\vspace{2pt}

\section{Related Work}
The overarching motivation for research on VQA is that of tackling a complex, open-world and multimodal task. These aspects are among the foundations required in general AI systems. While the task has attracted considerable attention over the past few years~\cite{teney2017visual,kafle2017visual}, its open-set and open-domain aspects have largely been overlooked. The common practice of training a model with end-to-end supervision using a fixed dataset is inherently limited. Our discussion focuses on the incorporation of additional knowledge and training signals into VQA models.

\paragraph{Answer embeddings for VQA.}
\label{subsec:related-vqa}
Most techniques to incorporate additional information into VQA models are based on representations of language, both of questions and of candidate answers. In~\cite{teney2016zero} pretrained word embeddings are used as bag-of-words representations of candidate answers, which are passed to the network as additional inputs, along with question and image features. In~\cite{teney2018tips} authors proposed to initialize the weights of the output classifier with pretrained answer embeddings. They used both a textual branch, initialized with GloVe vectors, and a visual one, initialized with visual features from images representing the candidate answers. In~\cite{hu2018learning}, the authors propose to learn two sets of embeddings, image-question vectors and answer embeddings. They optimize a projection of these two embeddings into a joint space where the distances between compatible pairs are minimized. Their experiments showed interestingly that the learned projections was transferable, to some extent, across datasets with different sets of possible answers.

Different from the methods cited above, our model forgoes the notion of a fixed answer set, and the output of the network is a location in a space representing answer semantics. The final prediction is still obtained by searching for the closest representation among answer candidates in this same space, but the formulation offers improved flexibility. This allows us to explore different distance measures in this semantic space. It also allows control over the contribution made by prior and task-specific data. Finally, it easily accommodates multiple representations of a same answer, thereby accounting for polysemy and context-dependent meaning of certain words and expressions.


\paragraph{Class embeddings for image classification.}
A related line of works use non-visual data to improve image classifiers. Techniques have been proposed to use unannotated text~\cite{frome2013devise}, knowledge graphs~\cite{xu2018fine} or hierarchical word databases~\cite{akata2015evaluation} to obtain meaningful class embeddings, which proved beneficial for fine-grained image classification. Our work applies similar ideas to the task of VQA, where the key challenge is to find embeddings semantically connecting both visual and textual modalities.

\section{Proposed Approach}

\begin{figure*}[t]
  \centering
  \includegraphics[width=0.85\linewidth]{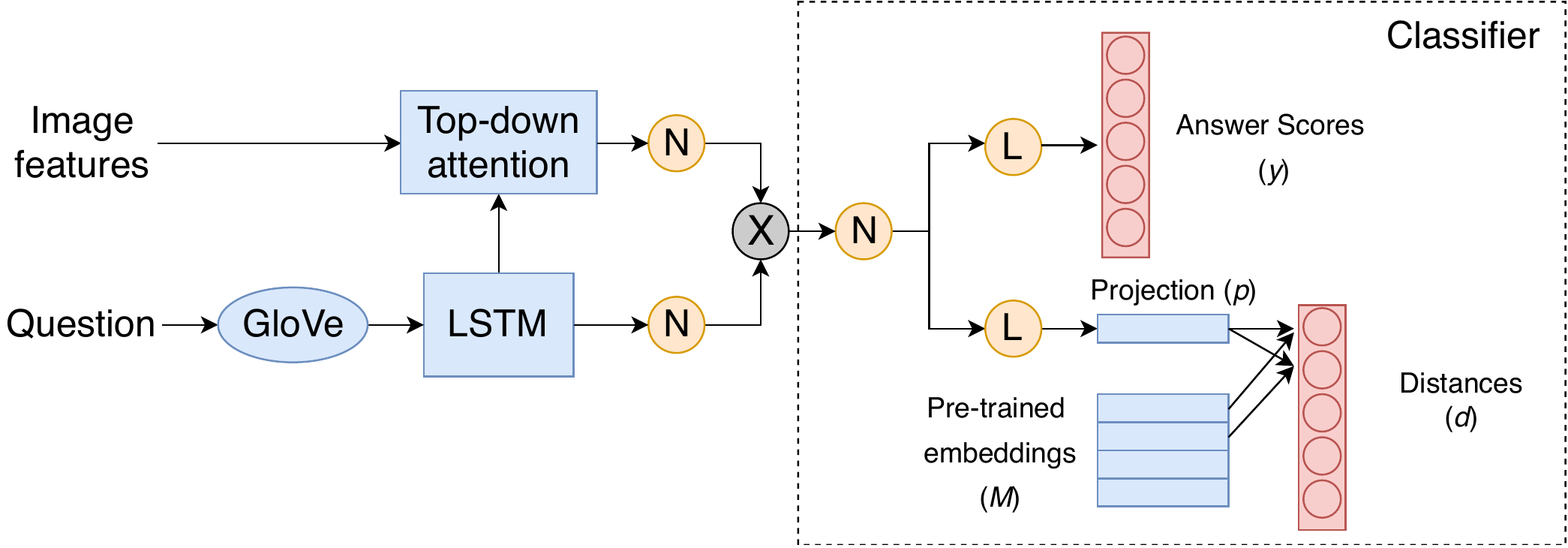}
  \caption{Our contributions apply to the classifier stage (dashed box) of a VQA model. We feed the fused image/question representation into two separate branches. (1)~In the upper branch, a traditional scoring model over predefined candidate answers. (2)~In the lower branch, a novel, learned projection to a semantic answer space. The resulting vector $p$ serves to measure pairwise distances ($d$) with pretrained representations of candidate answers ($M$). Nodes marked \textsf{N} denote non-linear layers, \textsf{L} linear layers, and \textsf{X} an element-wise product.} 
  \label{fig:model}
\end{figure*}

\label{subsec:architecture}
Our main idea is to extend VQA with a regression objective, where the model outputs a high-dimensional vector that represents the semantics of the answer. This is a shift from the traditional classification objective over predefined candidate answers. Our formulation will open the door to compositional and unbounded sets of answers, and the possibility of truly open-set prediction. Technically, our method concerns only the latter stage of a VQA model and is thus applicable to most existing ``joint embedding'' models, such as~\cite{antol2015vqa,zhou2015simple,gao2015you,saito2017dualnet}. In these models, the network produces a vector $\bx$ from the fusion of the image and  question representations (see \fig\ref{fig:model}). The traditional approach then feeds this to a classifier and obtain $\by=f_\theta(\bx)$, with $\by \in \mathbb{R}^\cardinality{A}$ being a vector of scores of length $A$, the cardinality of a predefined set of candidate answers.

\subsection{VQA as a Regression Task}
Our contribution is to learn a supplementary branch from $\bx$, which produces a projection $\bp=g_\psi(\bx)$, where $\psi$ are the parameters of the projection. The vector $\bp \in \mathbb{R}^\cardinality{P}$ is interpreted as a representation of the semantics of the predicted answer. The key to this simple approach is both in the objective used to train this branch, and in its use to select an actual textual answer, which we both describe below.

Note that the traditional classifier over $\bx$ can be interpreted as a special case of our formulation. The classifier $f_\theta(\cdot)$ typically includes a non-linear layer followed by a linear one. They can be interpreted as a non-linear projection followed by the computation of distances (dot products) with representations of answers. These representations then correspond to the rows of the weight matrix of the linear layer. In this view, our model is a generalization of the classical approach, with benefits of increased flexibility in the choice of the distance measure, of the optimization loss, and of the representations of candidate answers including their initial and/or frozen values.


\subsection{Training}
\label{subsec:training}
To evaluate the possibility of mutual benefits of the classification and regression objectives, our full model includes both branches on top of the fused representation $\bx$. Each of their respective outputs $\by$ and $\bp$ is fed into a specific loss. The whole network is trained by backpropagation of the gradient of the two losses through all the layers leading to $\bx$.


\paragraph{Classification loss.}
The output of the classification branch $\by$ goes through a standard logistic function $\sigma(\cdot)$ and binary cross entropy loss $L_{c}$. Denoting with $\hat{\ba} \in {0,1}^\cardinality{A}$ the one-hot (multi-hot) vector of the ground truth answer(s) of a specific training instance, we have    
\begin{equation}
  L_{c} = \sum_{i=1}^{A} -[\hat{a}_{i} * \log \sigma(y_{i}) + (1 - \hat{a}_{i}) * \log(1 - \sigma(y_{i}))]
\end{equation}
where $i$ indexes vector elements. The sum allows for multiple ground truth answers to a single training question.

\paragraph{Regression loss.}
The output of the additional regression branch produces the vector $\bp \in \mathbb{R}^\cardinality{P}$. It is interpreted as a location in a high-dimensional space that captures the semantics of the predicted answer. We store in a matrix $M_{\cardinality{A} \times \cardinality{P}}$ representations of $\cardinality{A}$ candidate answers in this space ($\cardinality{P}$-dimensional row vectors). These representations can be learned or initialized using prior knowledge, as described below. The objective of the regression branch is to produce a vector $\bp$ close to the representation of the ground truth answer, and distinct from those of incorrect ones. Using a metric $\dist(\cdot,\cdot)$, we compute all distances between $\bp$ and the rows of $M$, noted as $M_{i}$. We have
\begin{equation}
  \bd = [d_{1}, d_{2}, ..., d_\cardinality{A}] ~~~\textrm{with}~~ d_{i} = \dist(\bp, M_{i})~.
\end{equation}

We then define a hinge loss on these distances:
\begin{equation}
\resizebox{\columnwidth}{!}{$
  L_{p} = \sum_{i=1}^{\cardinality{A}} l_{i} ~~\textrm{with}~~
  l_{i} = \begin{cases}
        d_{i} & \text{if } \hat{a}_{i} = 1,\\
        \max\{ 0, \delta - d_{i} \} & \text{if } \hat{a}_{i} = 0.
      \end{cases}$}
\end{equation}
where $\delta$ is a scalar margin hyperparameter. Our overall optimization objective is the convex combination of the classification and regression losses:
\begin{equation}\label{eq:loss}
  L ~=~ \lambda \; L_{c} ~+~ (1-\lambda) \; L_{p} ~,
\end{equation}
where the scalar hyperparameter $\lambda$ balances the two objectives. By setting $\lambda=1$, the loss falls back to a unique traditional classification objective, which serves as our baseline.

\subsection{Predictions}
Due to the nature of existing datasets, answer prediction during test time do not differ from the training, since both train and test splits typically share common answer set. Our current experiments thus simply use the answers predicted by the network with the same combination of the classification and regression branches as the training objective. That is, the final predicted answer $a^\star \in [1...\cardinality{A}]$ is the one from the set of candidates with the combination of highest score and the lowest distance. Formally:
\begin{equation}\label{eq:prediction}
\resizebox{\columnwidth}{!}{$
  a^\star = \argmax_i \big(\lambda \,\softmax(\by) \,+\, (1\hspace{-.2em}-\hspace{-.2em}\lambda) \,\softmax(-\bd) \big)~.$}
\end{equation}

\subsection{Incorporating Prior Knowledge about Answers}
The matrix $M$ of the regression branch contains, in each of its rows, the representation of a candidate answer. $M$ can be treated and optimized as any other parameter of the network, but it can also be initialized with values that contain prior knowledge about answers. In particular, we experiment with GloVe embeddings~\cite{pennington2014glove} for single-word answers, and averaged (\ie as a bag-of-words) in the case of multi-word ones. The values of $M$ are further fine-tuned during training. Freezing them always proved inferior in our preliminary experiments (not reported).

As ablations of our model, we consider two other initialization schemes of $M$. They will serve to probe for the source of the gains of our model.
\vspace{-\topsep}
\vspace{-2pt}
\begin{itemize}
  \setlength\itemsep{1pt}
  \item \textit{Random.} We initialize $M$ with normally distributed random values, as would be any other weight matrix of the network.
  \item \textit{Shuffled GloVe.} We initialize $M$ with GloVe embeddings as described above, but subsequently shuffle its rows randomly, as in~\cite{teney2018tips}. The rows of $M$ are thus mismatched from their corresponding answers. This allows us to disentangle the anticipated benefits of using the semantic information carried in GloVe vectors, from the mere numerical effects of using them as initial values.
\end{itemize}


\section{Experiments}

\label{subsec:setting}
We performed an extensive evaluation to thoroughly validate the benefits of the proposed method, and understand the exact source of improvement. The overall conclusion is that the improvements indeed stem from the information brought in by the use of external data, rather than numerical artifacts or structural modifications to the network architecture.

Our contributions are implemented on top of the open-source Pythia framework~\cite{jiang2018pythia}, the winning entry of the 2018 VQA Challenge. The technique is however applicable to a wide range of current and future models. Pythia thus serves as the main baseline. We also evaluate the Pythia model where the weights of the output classifier are initialized with pretrained answer embeddings (noted `Pythia+GloVe') in the manner proposed by~\cite{teney2018tips}. We also compare our method to existing methods designed to inject prior knowledge in the model in the form of answer embeddings. Precisely, we consider the two variants of the ``factorized Probabilistic Model of Compatibility'' (fPMC) proposed by~\cite{hu2018learning}, using the code provided by the authors. All tested models use the same image features (those provided with the GQA dataset) and representations of question words (300-dimensional pretrained GloVe embeddings). Details are provided in \ap\ref*{appendix:baseline}.

The general hyperparameters of Pythia (batch size, learning rate, \etc) were chosen by grid search for best performance of the \emph{baseline} model (\ie without our contributions) on the GQA validation set. They were not modified once our contributions were added. This ensures a fair and challenging baseline. The distance function $\dist(\cdot,\cdot)$ is implemented as the Euclidean distance. This choice proved empirically superior, on the GQA validation set, to a dot product or a cosine similarity. The parameter $\lambda$ is set to $0.5$, unless otherwise noted. Every experiment was repeated with five different random seeds, and we report the average over the five runs. The ensembles use the average of the predicted scores/distances of several models trained with different random seeds, before taking the $\argmax$ of \eq\ref{eq:prediction}.


\begin{table*}[t!]
\ra{1.0}
\begin{center}
\resizebox{\textwidth}{!}{\begin{tabular}{@{}lccccccccccc@{}}
      \toprule
      & \multicolumn{3}{c}{GQA validation} &\phantom{abc}& \multicolumn{3}{c}{GQA test-dev} &\phantom{abc}& \multicolumn{3}{c}{GQA test}\\
      \cmidrule{2-4} \cmidrule{6-8} \cmidrule{10-12}
      & Binary & Open & All && Binary & Open & All && Binary & Open & All\\
      \midrule
      Blind LSTM & -- & -- & -- && -- & -- & -- && 61.90 & 22.69 & 41.07\\
      BUTD & -- & -- & -- && -- & -- & -- && 66.64 & 34.83 & 49.74\\
      MAC & -- & -- & -- && -- & -- & -- && 71.23 & 38.91 & 54.06\\
      LXMERT & -- & -- & -- && -- & -- & -- && 77.80 & 45.00 & 60.30\\
      NSM & -- & -- & -- && -- & -- & -- && \textbf{78.94} & \textbf{49.25} & \textbf{63.17}\\
      Pythia & 75.45 & 45.76 & 60.13 && 71.51 & 38.15 & 53.46 && -- & -- & --\\
      Pythia + GloVe & 74.91 & 45.77 & 59.87 && 71.36 & 37.94 & 53.28 && -- & -- & --\\
      fPMC(BUTD$\star$) & 69.85 & 42.28 & 55.62 && 64.80 & 35.40 & 48.90 && -- & -- & --\\
      fPMC(SAN$\star$) & 71.94 & 41.78 & 56.37 && 67.02 & 35.83 & 50.14 && -- & -- & --\\
      Ours + random & 75.15 & 46.33 & 60.27 && 70.67 & 38.14 & 53.08 && -- & -- & --\\
      Ours + shuffled GloVe & 76.17 & 46.53 & 60.87 && 71.80 & 38.48 & 53.78 && -- & -- & --\\
      Ours + GloVe & \textbf{76.93} & \textbf{46.99} & \textbf{61.48} && \textbf{72.19} & \textbf{39.31} & \textbf{54.40} && 71.35 & 40.07 & 54.73\\
      \midrule
      Ensemble: 5$\times$ Pythia & 77.24 & 48.41 & 62.36 && 73.43 & 39.85 & 55.26 && -- & -- & --\\
      Ensemble: 5$\times$ Ours + GloVe & \textbf{79.32} & \textbf{49.48} & \textbf{63.92} && \textbf{74.35} & \textbf{41.40} & \textbf{56.52} && -- & -- & --\\
      \bottomrule
\end{tabular}}
\caption{Accuracy (\%) on GQA. Our method shows clear improvements on both binary and open-ended questions.}
\label{Tab:overall-acc}
\vspace{-5pt}
\end{center}
\end{table*}

\begin{table*}[t!]
\ra{1.0}
\begin{center}
\renewcommand\tabcolsep{4pt}
\resizebox{\textwidth}{!}{\begin{tabular}{@{}lcccccccccc@{}}
      \toprule
      & \multicolumn{10}{c}{GQA test-dev}\\
      \cmidrule{2-11}
      & Choose & Compare & Logical & Query & Verify & Attribute & Category & Global & Object & Relation\\
      \midrule
      Pythia & 67.93 & 62.14 & \textbf{72.11} & 38.15 & 75.28 & 60.04 & 45.59 & 51.85 & 84.53 & 44.24\\
      Pythia + GloVe & 68.45 & \textbf{62.72} & 71.69 & 37.94 & 74.81 & 59.40 & 44.72 & 54.14 & 84.78 & 44.51\\
      fPMC(BUTD$\star$) & 60.39 & 57.83 & 63.36 & 35.40 & 69.99 & 51.17 & 42.90 & 51.97 & 81.95 & 43.04\\
      fPMC(SAN$\star$) & 64.80 & 61.53 & 65.62 & 35.83 & 70.69 & 53.80 & 43.69 & 54.01 & 80.95 & 43.32\\
      Ours + random & 67.42 & 62.07 & 70.87 & 38.14 & 74.40 & 58.53 & 45.01 & \textbf{55.42} & 84.68 & 44.79\\
      Ours + shuffled GloVe & 70.88 & 62.14 & 71.38 & 38.48 & 75.14 & 59.65 & 45.36 & 54.14 & 84.60 & 45.33\\
      Ours + GloVe & \textbf{71.12} & 62.21 & 71.14 & \textbf{39.31} & \textbf{76.17} & \textbf{60.28} & \textbf{46.04} & 53.88 & \textbf{85.01} & \textbf{46.01}\\
      \midrule
      Ensemble: 5$\times$ Pythia & 70.95 & 63.33 & \textbf{73.54} & 39.85 & 77.22 & 61.84 & 47.87 & 52.23 & \textbf{86.50} & 45.95\\
      Ensemble: 5$\times$ Ours + GloVe & \textbf{75.11} & \textbf{64.18} & 73.04 & \textbf{41.40} & \textbf{77.66} & \textbf{63.00} & \textbf{48.30} & \textbf{53.50} & 86.25 & \textbf{47.70}\\
      \bottomrule
\end{tabular}}
\caption{Accuracy (\%) over question types on GQA test-dev. Our contributions bring clear improvements on most question types, with the highest gain on the \textit{choose} category.}
\label{Tab:quest-test-acc}
\vspace{-5pt}
\end{center}
\end{table*}

For evaluation of our approach we use the GQA dataset~\cite{hudson2018gqa} as it provides the most comprehensive suite of metrics and cleanest data of current VQA datasets. However, we do not aim to build a data-specific solution, so our model does not utilize scene graphs and functional programs included in the dataset.

\subsection{Quantitative Results}
Our main results on the GQA dataset are provided in \tab\ref{Tab:overall-acc}. Looking at the overall accuracy, our model clearly outperforms all baselines and ablations. The same observations can be drawn on both the binary and open-ended questions. The trend is also confirmed when evaluating an ensemble of our model, versus a similar ensemble of the Pythia baseline. The fPMC model~\cite{hu2018learning} obtains the lowest results, including our modified version fPMC(BUTD$\star$) (details in \ap\ref*{appendix:baseline}), which indicates its lack of adaptivity to complex feature representation methods. The fPMC model was initially tested only on the very noisy VQA v2 dataset, and a possible reason for its weak performance on GQA is the narrower answer set. A surprising outcome is that Pythia with pretrained classifier (`Pythia+GloVe') shows worse accuracy results than the baseline. This occurs mostly due to overfitting of the pre-initialized classifier to the most common answers in the training set, as observed by the reduced accuracy on both the validation (see \tab\ref*{Tab:quest-val-acc} in \ap\ref*{appendix:results}) and test-dev sets. Unlike the other described architectures, our model exploits the additional information contained in the representations of answers in an effective way, increasing performance without overfitting.  


\subsection{Comparison with Existing Models}
We compare our model with existing methods reported in~\cite{hudson2018gqa} and several recent state-of-the-art (see \tab\ref{Tab:overall-acc}). We report the performance of the blind LSTM, the bottom-up top-down attention model~\cite{anderson2018bottom}, MAC~\cite{hudson2018compositional}, LXMERT~\cite{tan2019lxmert} and the neural state machine (NSM)~\cite{hudson2019learning}. Our model shows better results than all the baselines, and in spite of a much simpler architecture, it notably surpasses the MAC model. However, the newest methods LXMERT and NSM show higher performance which is not surprising. LXMERT model explores more sophisticated technique of image and language representation and is pretrained on a significantly larger amount of data. NSM implements compositional approach and performs explicit multi-step reasoning. Differently, our approach focuses on the output stage of VQA model, thus the contributions of this article are expected to be applicable to these models.

\begin{table}[t!]
\ra{1.0}
\begin{center}
\resizebox{\columnwidth}{!}{\begin{tabular}{@{}lcccc@{}}
      \toprule
      & \multicolumn{4}{c}{GQA validation}\\
      \cmidrule{2-5}
      & V & P & D & C\\
      \midrule
      Pythia & 95.07 & 91.39 & \textbf{3.93} & 83.12\\
      Pythia + GloVe & 95.13 & 91.40 & 4.07 & 82.68\\
      fPMC(BUTD$\star$) & 94.99 & 90.91 & 6.20 & 76.53\\
      fPMC(SAN$\star$) & 95.11 & \textbf{91.62} & 5.66 & 78.67\\
      Ours + random & 95.07 & 91.53 & 4.26 & 83.00\\
      Ours + shuffled GloVe & 95.14 & 91.48 & 4.01 & 83.37\\
      Ours + GloVe & \textbf{95.16} & 91.55 & 4.01 & \textbf{84.57}\\
      \midrule
      Ensemble: 5$\times$ Pythia & 95.17 & 91.90 & 4.78 & 85.27\\
      Ensemble: 5$\times$ Ours + GloVe & \textbf{95.25} & \textbf{92.06} & \textbf{4.56} & \textbf{87.33}\\
      \bottomrule
\end{tabular}}
\caption{Results on additional metrics: \underline{V}alidity, \underline{P}lausibility, \underline{D}istribution (lower is better), and \underline{C}onsistency. Our model noticeably improves in consistency over the baseline. It ranks slightly worse on the  distribution metric (see discussion in text).}
\label{Tab:other-metrics}
\end{center}
\end{table}

\subsection{In-Depth Analysis}
\begin{figure}[thp]
  \centering
  \includegraphics[width=1\linewidth]{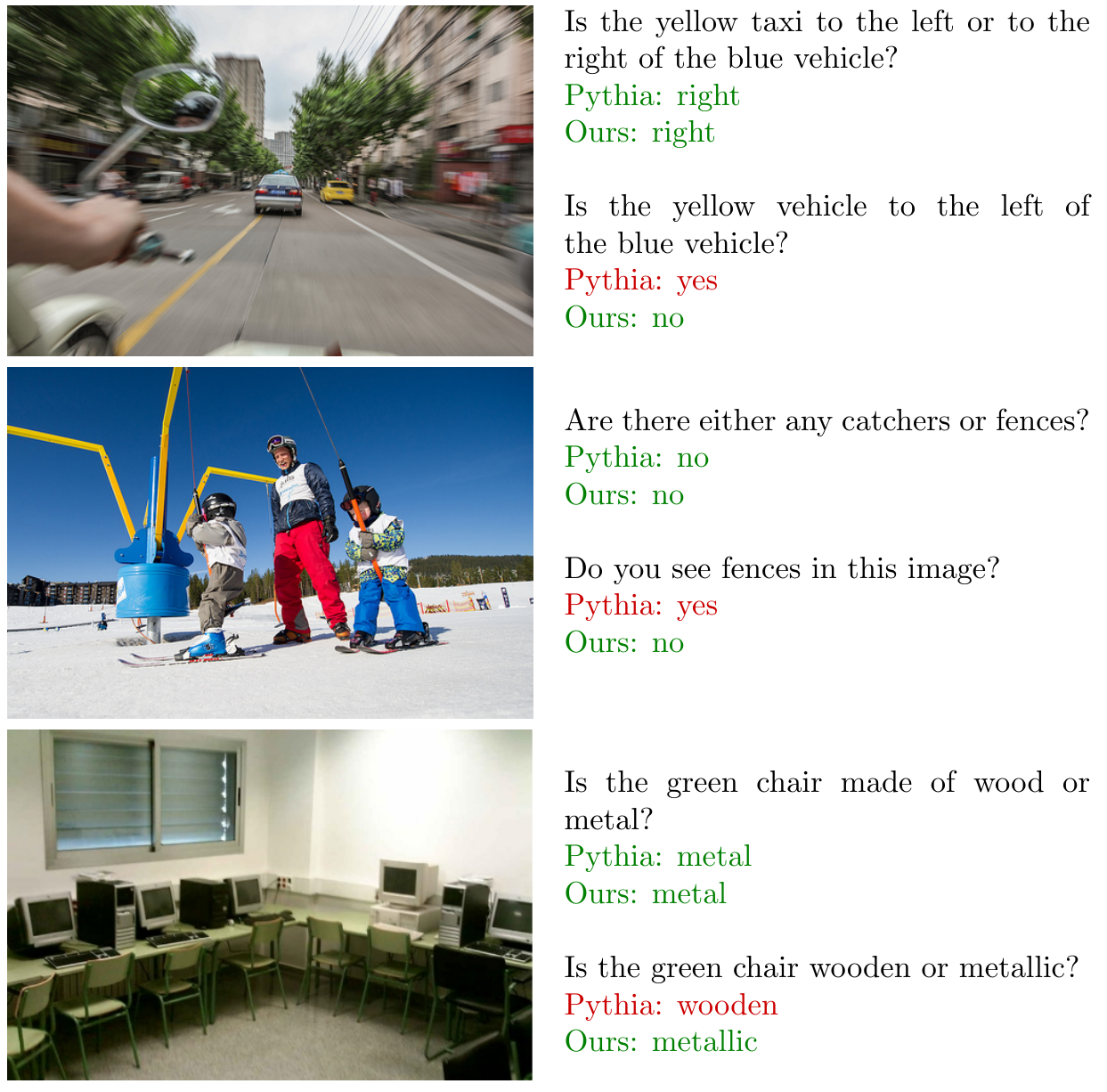}
  \caption{Qualitative examples from the GQA dataset, with predictions of our model and of the Pythia baseline. We show pairs of questions about a same image where the first entails the second (this information is never provided to the model during training nor testing). Our model improves in consistency over the baseline, producing pairs of answers compatible with one another.}
  \label{fig:consistency}
\end{figure}

We report the detailed metrics of the GQA dataset in \tab\ref{Tab:other-metrics}. The first observation is that a similar ranking of methods and ablations can be drawn from most of the metrics. This stability further confirms the benefits of the proposed method. The improvements on these advanced metrics also indicate benefits beyond the sole increase in accuracy. The \textit{validity} and \textit{plausibility} scores, in particular, which are noticeably higher, indicate a generally more robust model. The higher \textit{consistency} score implies that the answers produced over related questions are compatible with one another (see \fig\ref{fig:consistency}). The only metric on which our model falls below the baseline is the \textit{answer distribution}. It indicates that the model occasionally favors one answer over most others. We explain this by the fact that some answers are not assigned appropriate initial representations (see the rest of the discussion below). We also look at the accuracy metric broken down by question categories (\tab\ref{Tab:quest-test-acc}). We observe no significant drop in accuracy for any type, and the highest improvements occur on the \textit{choose}, \textit{query}, \textit{attribute}, and \textit{relational} questions.

\begin{figure}[t!]
  \centering
  \includegraphics[width=0.9\linewidth]{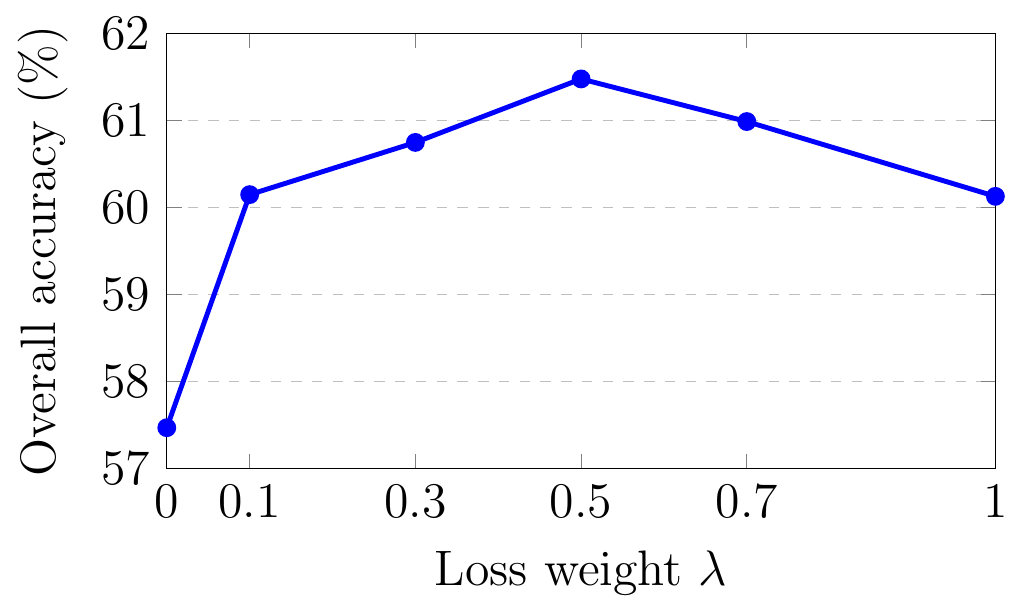}
  \vspace{-2pt}
  \caption{The performance of our model varies smoothly with the relative weight of the classification and regression losses (\eq~\ref{eq:loss}). The value $\lambda$=$1$ corresponds to a traditional classification-only baseline, while the optimal value $\lambda$=$0.5$ corresponds to an even contribution of the two losses.}
  \label{fig:lambda_plot}
  \vspace{-3pt}
\end{figure}

The ablations of our method (\textit{Ours+random} and \textit{Ours+shuffled GloVe}) are important to determine whether the source of improvements is in the architecture of our model (the additional output branch and loss), in numerical effects from the initialization of the matrix $M$ with values from GloVe vectors, or in the actual information conveyed in the GloVe vectors. The ablation with random initial values is essentially similar to the Pythia baseline, which shows no significant effect from the architecture alone. Surprisingly, the \textit{shuffled GloVe} ablation brings some improvement, which we explain by two factors. First, since the values of $M$ are further fine-tuned with the rest of model, they can still incorporate useful information from the task-specific supervision even if the initial values do not contain relevant semantic information. Second, some answers may actually benefit from the ``wrong'' initialization: we have determined that the absolute values of the representations of answers do not play the most significant role, but that their mutual relations are what encodes the critical information. This shows up in particular with pairs of antonym answers such as \textit{yes}/\textit{no} or \textit{left}/\textit{right}. The GloVe embeddings of these pairs are usually similar, whereas the VQA task-specific supervision tends to push their representations apart. The \textit{shuffled} initialization can thus prove better than the ``correct'' one for some cases. This can also be observed on the high accuracy of the \textit{shuffled} ablation on the \textit{choose} category of questions which do specifically contain this type of antonym answers (see \tab\ref{Tab:qtypes}). Despite these effects, the full model still performs clearly better than the ablations, indicating an overall benefit from the information conveyed in the GloVe representations of answers.

Since our architecture is trained to minimize a sum of two losses (classification and regression), we sought to evaluate their possible mutual benefit by varying their relative weight ($\lambda$ in \eq\ref{eq:loss}). A Value of $\lambda$=$0$ corresponds to the regression loss alone, and $\lambda$=$1$ to the baseline using the traditional classification loss alone. Interestingly, a balanced value of $0.5$ leads to the highest accuracy (\fig\ref{fig:lambda_plot}), demonstrating that they are indeed complementary.

\begin{figure}[t!]
  \centering
  \includegraphics[width=1\linewidth]{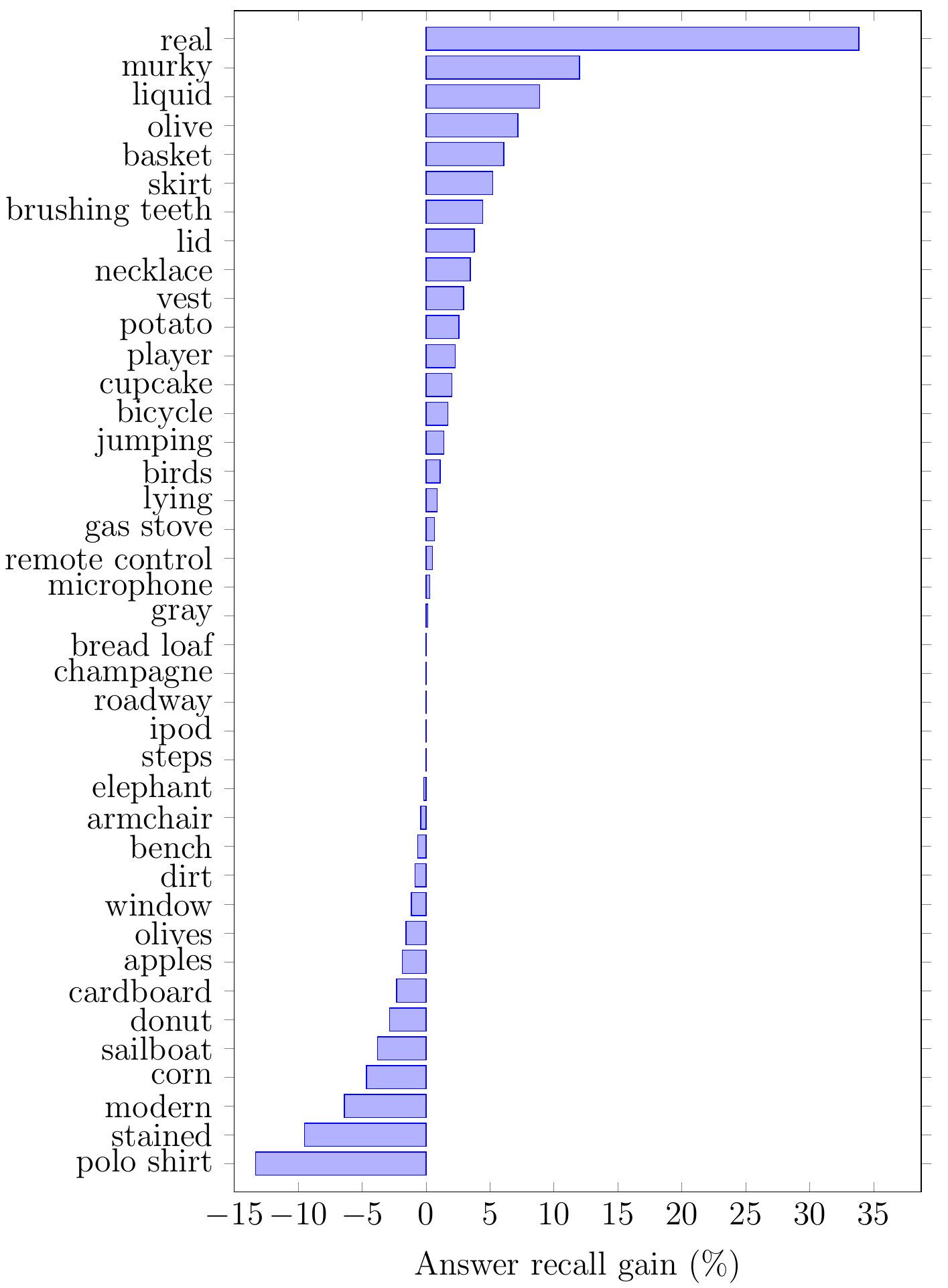}
  \caption{Absolute gain in answer recall of our model over the Pythia baseline (positive is an improvement). We report an even subset of answers (every $25^{\text{\tiny th}}$ one in descending recall gain). See the text for a discussion.}
  \label{fig:answer_recall}
\end{figure}

To obtain deeper insights into the additional knowledge that is actually most beneficial, we examined the improvements of our model over the Pythia baseline on individual answers. We report, in \fig\ref{fig:answer_recall}, the change in answer recall for a random selection of answers. We define the answer recall as, for an answer candidate $\hat{a}$, the ratio of questions with $\hat{a}$ as ground truth that are correctly answered by the model. The recall of most answers improves, but it stays similar or even degrades on some others. We investigated the possible reasons. A degradation is presumably related to less relevant initial representations of the corresponding answer. To assess this, we examined the closest other answers in the space of pretrained GloVe vectors. Most answers with a negative gain in answer recall have neighbors with no semantic or syntactic connections. For instance, the three closest neighbors to \textit{modern} are $\{$\textit{under}, \textit{rooftop}, \textit{visitor}$\}$. Answers with a high recall improvement, on the contrary, tend to have semantically related neighbors. For example, \textit{basket} has the closest neighbors $\{$\textit{baskets}, \textit{cane}, \textit{sack}$\}$. These observations further support the claim that mutual relations between representations of answers are the major way in which the network stores and uses semantic information.

\subsection{Prediction of Novel Answers}
\label{sec:novel-ans}
Our model trained with the regression objective can predict answers at test time that are outside the predefined set of candidates used for training (\ie open-set prediction a.k.a. zero-shot VQA~\cite{teney2016zero}). This is achieved by replacing the matrix $M$ with new answers and setting $\lambda$ to 0 at test time. To evaluate this setting, we use ConceptNet embeddings~\cite{speer2017conceptnet}, which are designed to common sense knowledge.
We use the VQA v2 dataset~\cite{goyal2017making} since it features a more diverse set of answers than GQA. We use splits with disjoint sets of answers at training and test time (see details in \ap\ref*{appendix:data}). In this setting, our model achieves an accuracy of 27\% on the test set, while fPMC model, which also has tools for out-of-vocabulary prediction, obtains about 15\% accuracy. Given that test questions feature exclusively answers never seen during training, this clearly demonstrates a capability for predictions beyond the scope of the training set. However, the performance on novel answers is highly depended on the used answer representations.
Embeddings like GloVe and ConceptNet carry only limited, mostly linguistic information, which is insufficient for the full scope of their use in VQA.
We discovered that embeddings learned end-to-end in a VQA model capture visual co-occurrences (\fig\ref*{fig:tsne} in \ap\ref*{appendix:learned}). This implies that additional knowledge extracted from visual data (\eg as \cite{noh2019transfer}) should be a useful complement to boost out-of-vocabulary performance. The combination of multiple types of pretrained embeddings is a promising avenue for future work.



\section{Conclusions}
In this paper, we reformulated VQA as a multitask problem, which allowed us to exploit prior semantic knowledge about answers. We demonstrated that GloVe word embeddings carry information about typical answers that is relevant to the task. In contrast to existing methods for incorporating additional data into VQA models, our technique is both simple and effective, and allows to tune the reliance of the model on general prior knowledge, and learned task-specific information. We evaluated our technique on the GQA dataset and obtained consistent improvement in accuracy in the majority of question categories. The extensive set of metrics also allowed identifying benefits in robustness and consistency of the model across related questions.

The fundamental idea in this paper of including a regression task as part of VQA has implications that go beyond what could be demonstrated with existing datasets.
This formulation opens the door to the generation of compositional multi-word answers, and to open-set prediction, that is, predicting answers beyond the set of candidate answers predefined at training time.

{\small
\bibliographystyle{ieee_fullname}
\bibliography{egbib}
}

\clearpage

\appendix

\section*{Visual Question Answering with Prior Class Semantics}

\vspace{2pt}
\section*{Appendices}
\vspace{12pt}

\section{Baseline Methods}
\label{appendix:baseline}
\paragraph{Pythia}
Our baseline model is the Pythia implementation~\cite{jiang2018pythia} of the classical joint embedding model~\cite{teney2017visual,kafle2017visual}. Pythia uses object image features extracted with the pretrained Faster R-CNN model provided with the GQA dataset. On the language side, words are represented with word embeddings initialized with pretrained GloVe vectors, followed by an LSTM to produce a vector representation of the whole question. A question-guided top-down attention is applied on image features to identify relevant image regions. The image and question features are passed through non-linear layers and finally combined with an element-wise multiplication. The final classifier comprises a non-linear layer and a linear one, which produces a score for each candidate answer. All non-linear layers throughout the network use weight normalization~\cite{salimans2016weight} and ReLU activations.

Pythia serves as reference for evaluation, and as the bare model on which to build our contributions. This choice is justified by a few reasons. It is a high-performing open-source implementation that still outperforms many others on the VQA v2 dataset~\cite{goyal2017making}. This provides us with a strong --~and thus challenging~-- starting point to demonstrate the proposed method. Moreover, the implementation of Pythia is modular and easily allows one to separate, replace, and compare the various blocks of the model. In our case, this enables us to focus specifically on the classification part of the model, leaving the rest unchanged.

\paragraph{Pythia with pretrained classifier}
We compare our method to the Pythia model, in which the output classifier is initialized with pretrained answer embeddings. As discussed in the related works section, this is a reasonable approach to embed semantic information about candidate answers within the model. Following a procedure similar to~\cite{teney2018tips}, we collect 300-dimensional GloVe embeddings for all words in the answer vocabulary (substituting unknown words with zero vectors). We represent each answer directly by its matching word embedding, or, in the case of multi-word answers, by the average embedding of the constituent words. Next, we design the classifier block of the model as follows: one non-linear layer with output dimension equal to the dimensionality of used GloVe embeddings followed by a linear layer with a weight matrix $w \in \mathbb{R}^{300 \times A}$. Each row of $w$ thus contains the vector corresponding to one specific answer. Besides the non-random initialization of $w$, the only distinction with the original Pythia model is that the output dimension of the non-linear layer is reduced from 5000 to 300 to match the dimensionality of the GloVe vectors.

\paragraph{Factorized probabilistic model of compatibility}
We also compare the proposed approach to the factorized Probabilistic Model of Compatibility (fPMC)~\cite{hu2018learning}. In this architecture, a joint image-question embedding is learned alongside the answer embedding, and the model is trained to increase the likelihood of the correct answer. We performed all experiments with the following two variants proposed by the authors.
\vspace{-\topsep}
\begin{itemize}
  \setlength\itemsep{1pt}
  \item fPMC(SAN$\star$) model, described in the original paper, that utilizes stacked attention network~\cite{yang2016stacked} together with bidirectional LSTM and spatial image features extracted with ResNet-152. For obtaining answer embeddings, the model exploits two-layer bidirectional LSTM over GloVe vectors. We used the code provided by the authors of the paper and made the adjustments only required to make it compatible with GQA dataset.
  \item fPMC(BUTD$\star$) model is our modification of fPMC(SAN$\star$) where we used the ``bottom-up and top-down attention''~\cite{anderson2018bottom} model with object image features for parameterizing the joint embedding in the same way as all the other models used in our experiments. We were thus able to explicitly evaluate the approach of learning aligned answer embeddings independently from the impact of different feature initializations.
\end{itemize}

\section{Implementation of the Proposed Method}
\label{appendix:implementation}
The proposed method builds directly on the open source Pythia implementation\footnote{https://github.com/facebookresearch/pythia/tree/0.1}, which uses PyTorch. Our model is trained for 20,000 iterations with a batch size of 512 and AdaMax optimizer~\cite{kingma2014adam}. We adopted a warm-up learning schedule strategy from~\cite{jiang2018pythia} and tuned it to the current setup. Specifically, the starting learning rate of 0.002 is linearly growing up to 0.1 during first 1000 iterations and then decreased by a factor of 0.1 at 11,000, 13,000 and 15,000 iterations. Importantly, these hyperparameters were selected for best performance of the \textbf{baseline} model on the validation set of the GQA dataset, thus avoiding any unfair advantage for our contributions. The values of the regression loss margin ($\delta=1$) and of the loss weight ($\lambda=0.5$) were determined by grid search for best overall accuracy on the GQA validation set.

For VQA v2 dataset the only difference in hyperparameters is the learning schedule. The model is trained for 12000 iterations with learning rate decreasing at 5000, 7000, 9000 and 11,000 iterations, following the original Pythia implementation. We also found it beneficial to apply L2 normalization after projection layer. In out-of-vocabulary experimental setting we used a subset of VQA v2 data, so the parameter were adjusted to fit the smaller dataset. Specifically, we reduced the batch size to 128 and increased the number of iterations to 30,000 with decreasing steps at 12,000, 17,000, 22,000, and 25,000 iterations.

\section{Datasets}
\label{appendix:data}
\paragraph{GQA}
The dataset was designed as benchmark for compositional question answering over real-world images. The authors proposed new metrics for a detailed assessment of a model's performance:
\vspace{-\topsep}
\begin{itemize}
  \setlength\itemsep{1pt}
  \item \textit{Validity} measures whether the predicted answer fits the scope of the question (\eg a number for a counting question).
  \item \textit{Plausibility} checks that the answer is semantically reasonable, defined as occurring at least once with the given question in the whole dataset.
  \item \textit{Distribution} is the $\chi^2$ distance between the distributions of predicted and ground truth answers over groups of questions. A lower value means a better ability to predict less frequent answers.
  \item \textit{Consistency} measures the agreement between answers to pairs of questions about a same image where one entails the other.
  \item \textit{Grounding} is used for evaluation of attention-based models and is not tested in our study since an attention is not the focus of this research.    
\end{itemize}
The dataset also assigns test questions to categories (\tab\ref{Tab:qtypes}), across which the accuracy can be measured separately (as done in \tab\ref*{Tab:quest-test-acc}).


\begin{table}[]
\ra{1.0}
\begin{center}
\resizebox{\columnwidth}{!}{\begin{tabular}{@{}ll@{}}
      \toprule
      Type & Example\\
      \midrule
      Choose & Is it an indoors or outdoors scene~?\\
      Compare & Are all these animals of the same type~?\\
      Logical & Are there nuts or vegetables~?\\
      Query & What is this bird called~?\\
      Verify & Is there a cat that is not white~?\\
      Attribute & What is the color of the fence made of metal~?\\
      Category & What piece of furniture is not small~?\\
      Global & Which place is it~?\\
      Object & Is there a train in the picture~?\\
      Relation & What is the vegetable on top of the pizza~?\\
      \bottomrule
\end{tabular}}
\caption{Examples of each question type of the GQA dataset.}
\label{Tab:qtypes}
\end{center}
\end{table}


\paragraph{VQA v2 with Novel Answers}
For testing the out-of-vocabulary answer prediction, we created a subset of VQA v2 dataset. We used the original training and validation splits as our new training and test splits respectively. In each of them, we filtered the questions according to the following rules:
\begin{inparaenum}[1)]
  \item every ground truth answer has a corresponding ConceptNet embedding (exact match),
  \item every ground truth answer consists of one word only (\eg discarding \textit{black and white} or \textit{don't know}),
  \item every ground truth answer must occur in the original dataset between 5 and 500 times (thus discarding very rare and extremely frequent answers such as \textit{yes} and \textit{no}),
  \item the sets of ground truth answers in the training and test splits do not intersect.
\end{inparaenum}  
With this procedure, we obtain 91,255 training questions with 6,928 possible answers, and 13,367 test questions with another 1,187 answers.

\section{Additional Results}
\label{appendix:results}

\subsection{GQA}
We provide in \tab\ref{Tab:quest-val-acc} the performance on the GQA validation set, matching the experiments provided in \tab\ref*{Tab:quest-test-acc} of the main paper.

\begin{table*}
\ra{1.0}
\begin{center}
\resizebox{\textwidth}{!}{\begin{tabular}{@{}lcccccccccc@{}}
      \toprule
      & \multicolumn{10}{c}{GQA validation}\\
      \cmidrule{2-11}
      & Choose & Compare & Logical & Query & Verify & Attribute & Category & Global & Object & Relation\\
      \midrule
      Pythia & 70.64 & \textbf{67.57} & \textbf{81.66} & 45.76 & 75.86 & 64.43 & 56.58 & 64.81 & 83.69 & 51.42\\
      Pythia + GloVe & 69.94 & 67.11 & 81.28 & 45.77 & 75.30 & 63.84 & 57.19 & 65.20 & 83.44 & 51.22\\
      fPMC(BUTD$\star$) & 65.05 & 60.54 & 75.93 & 42.28 & 70.54 & 55.32 & 53.91 & 62.61 & 80.00 & 49.45\\
      fPMC(SAN$\star$) & 69.72 & 64.04 & 77.38 & 41.78 & 71.25 & 56.28 & 55.20 & 63.99 & 80.38 & 50.04\\
      Ours + random & 71.08 & 67.02 & 81.13 & 46.33 & 75.28 & 63.63 & 56.52 & 65.66 & 83.36 & 52.32\\
      Ours + shuffled GloVe & 74.27 & 66.92 & 81.31 & 46.53 & 75.68 & 64.88 & 56.89 & \textbf{65.74} & 83.54 & 52.64\\
      Ours + GloVe & \textbf{75.36} & 67.33 & 81.60 & \textbf{46.99} & \textbf{76.58} & \textbf{66.21} & \textbf{57.23} & 65.58 & \textbf{83.71} & \textbf{52.94}\\
      \midrule
      Ensemble: 5$\times$ Pythia & 73.65 & \textbf{69.40} & 82.92 & 48.41 & 77.23 & 67.48 & 58.77 & 65.90 & 84.71 & 53.48\\
      Ensemble: 5$\times$ Ours + GloVe & \textbf{79.28} & 69.33 & \textbf{83.08} & \textbf{49.48} & \textbf{78.65} & \textbf{69.38} & \textbf{58.83} & \textbf{66.54} & \textbf{85.09} & \textbf{55.37}\\
      \bottomrule
\end{tabular}}
\caption{Performance for different question types (accuracy in \%) on the GQA validation split.}
\label{Tab:quest-val-acc}
\end{center}
\end{table*}

\subsection{VQA v2}
\label{appendix:vqaresults}
\begin{table*}[t!]
\ra{1.0}
\begin{center}
\resizebox{0.9\textwidth}{!}{\begin{tabular}{@{}lccccccccc@{}}
      \toprule
      & \multicolumn{4}{c}{VQA v2 validation} &\phantom{abc}& \multicolumn{4}{c}{VQA v2 test-dev}\\
      \cmidrule{2-5} \cmidrule{7-9}
      & Yes/No & Number & Other & All && Yes/No & Number & Other & All\\
      \midrule
      Pythia  & \textbf{83.11} & 44.50 & 56.86 & 65.11 && \textbf{83.42} & 45.53 & 57.13 & 66.64 \\
      Ours + GloVe  & 82.90 & \textbf{44.68} & 56.93 & 65.08 && 83.33 & 45.46 & 57.18 & 66.62 \\
      Ours + GloVe (w/o fine-tuning) & 82.87 & 44.55 & \textbf{57.19} & \textbf{65.18} && 83.20 & \textbf{45.54} & \textbf{57.53} & \textbf{66.74} \\
      \bottomrule
\end{tabular}}
\caption{Accuracy (\%) on VQA v2 validation and test-dev sets. Our model with fixed (not fine-tuned) GloVe embeddings shows the highest results on the category \textit{other} and on all questions overall for both splits.}
\label{Tab:vqa-acc}
\end{center}
\end{table*}

We present experiments on the VQA v2 dataset in \tab\ref{Tab:vqa-acc}. Contrary to our results on GQA, we observe no significant difference compared to the baseline. We attribute this to the nature of the dataset.
In VQA v2, a large fraction of the questions (over 37\%) are to be answered with \textit{yes} or \textit{no}, and another 13\% with a number.
Our approach, which focuses on the representation of answer semantics, is already expected to have no influence on this large part of the dataset.
Moreover, numbers in VQA v2 are used not only for counting questions, but also to refer to abstract concepts, as in questions like \textit{How old is animal~?}, \textit{What time is the clock showing~?}, or \textit{What is the size of the TV~?}. It would certainly be difficult to infer a single representation of numbers that would encompass such a variety of concepts.

An additional challenge with VQA v2 is that most questions have multiple ground truth answers that are actually synonyms. Other times, annotation noise means that multiple answers with contradictory meanings are marked as correct. For example, a question \textit{Is the dog male or female~?} has both \textit{male} and \textit{female} answers in the annotation. In our model, all ground truth answers contribute equally to the projection loss, meaning that noisy or incorrect answer labels can push the learned projection in wrong directions. This issue could be mitigated by introducing instance-specific weights in the projection loss. This is an interesting avenue for future work.

Overall, our approach still has a positive impact on VQA v2 for out-of-vocabulary prediction (see \sect\ref*{sec:novel-ans}). And importantly, the above issues did not incur a decrease in performance compared to the baseline model.

\subsection{Learned Representations}
\label{appendix:learned}
\begin{figure*}
\centering
\subfigure[GloVe embeddings.]{\includegraphics[width=0.45\linewidth]{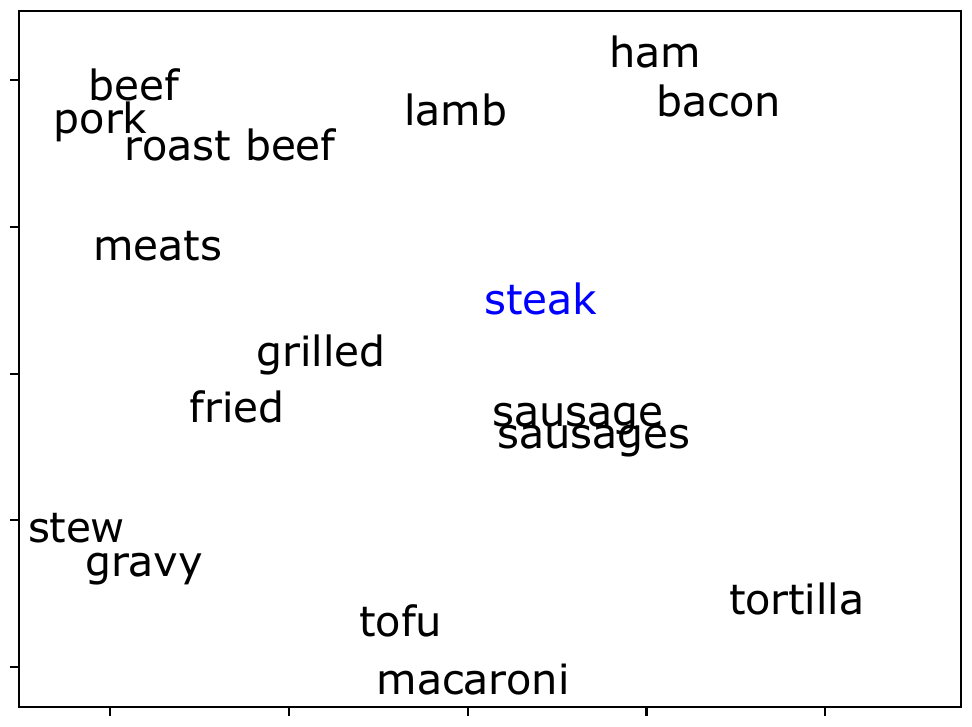}}
\subfigure[Learned answer representations.]{\includegraphics[width=0.45\linewidth]{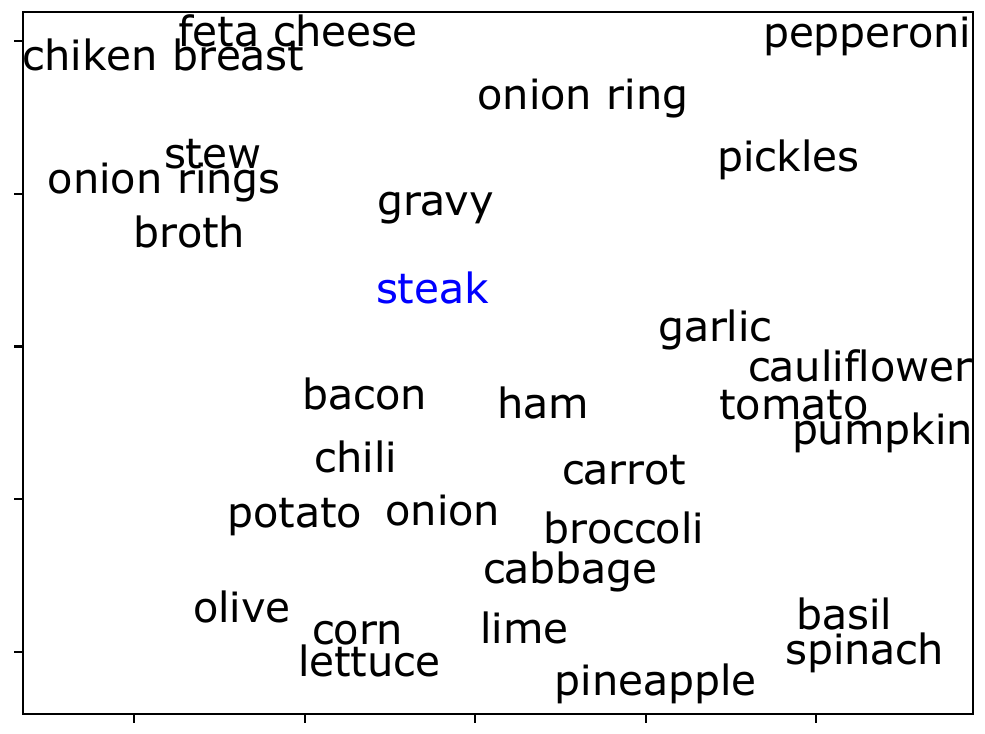}}
\subfigure[Example images from the GQA dataset.]{\includegraphics[width=0.45\linewidth]{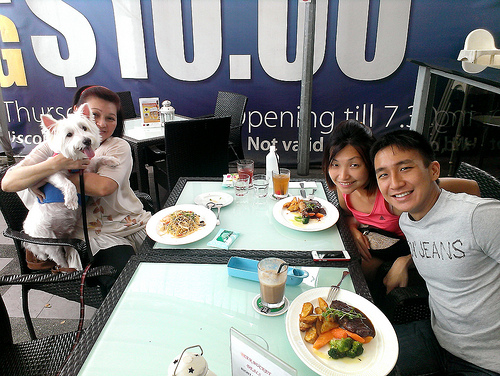}~\includegraphics[width=0.45\linewidth]{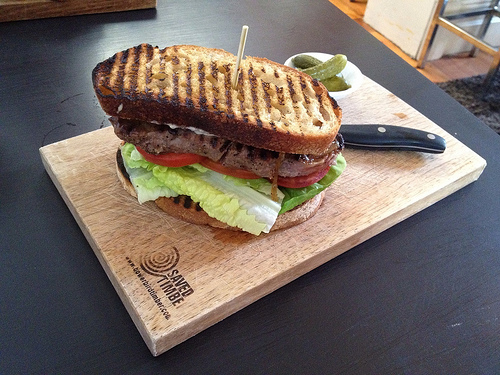}}
\caption{Examples of t-SNE projections in 2D of (a)~initial and (b)~fine-tuned representations of answers. The proximity of the learned representations better captures typical co-occurrences of the corresponding concepts in images from the dataset~(c).}
\label{fig:tsne}
\end{figure*}

We use t-SNE projections to visualize and compare off-the-shelf GloVe embeddings of candidate answers, which we use as prior knowledge to initialize the representations, with these representations after fine-tuning within our VQA model. As expected, the GloVe embeddings carry the kind of semantic similarity that emerges from co-occurrence of words in natural language. In the fine-tuned representations, we rather observe that the proximity of representations captures common co-occurrences of concepts in a same image, such that they are plausible answers to possible questions about this image. For example, the word \textit{steak} is projected close to the words \{\textit{potato}, \textit{carrot}, \textit{broccoli}, \textit{tomato}, \textit{pickles}\} (\fig\ref{fig:tsne}~(b)). We indeed observe co-occurrence of these objects in images from the GQA dataset (\fig\ref{fig:tsne}~(c)).

\subsection{Prediction of Novel Answers}

\begin{figure*}[htp]
  \centering
  \includegraphics[width=\textwidth]{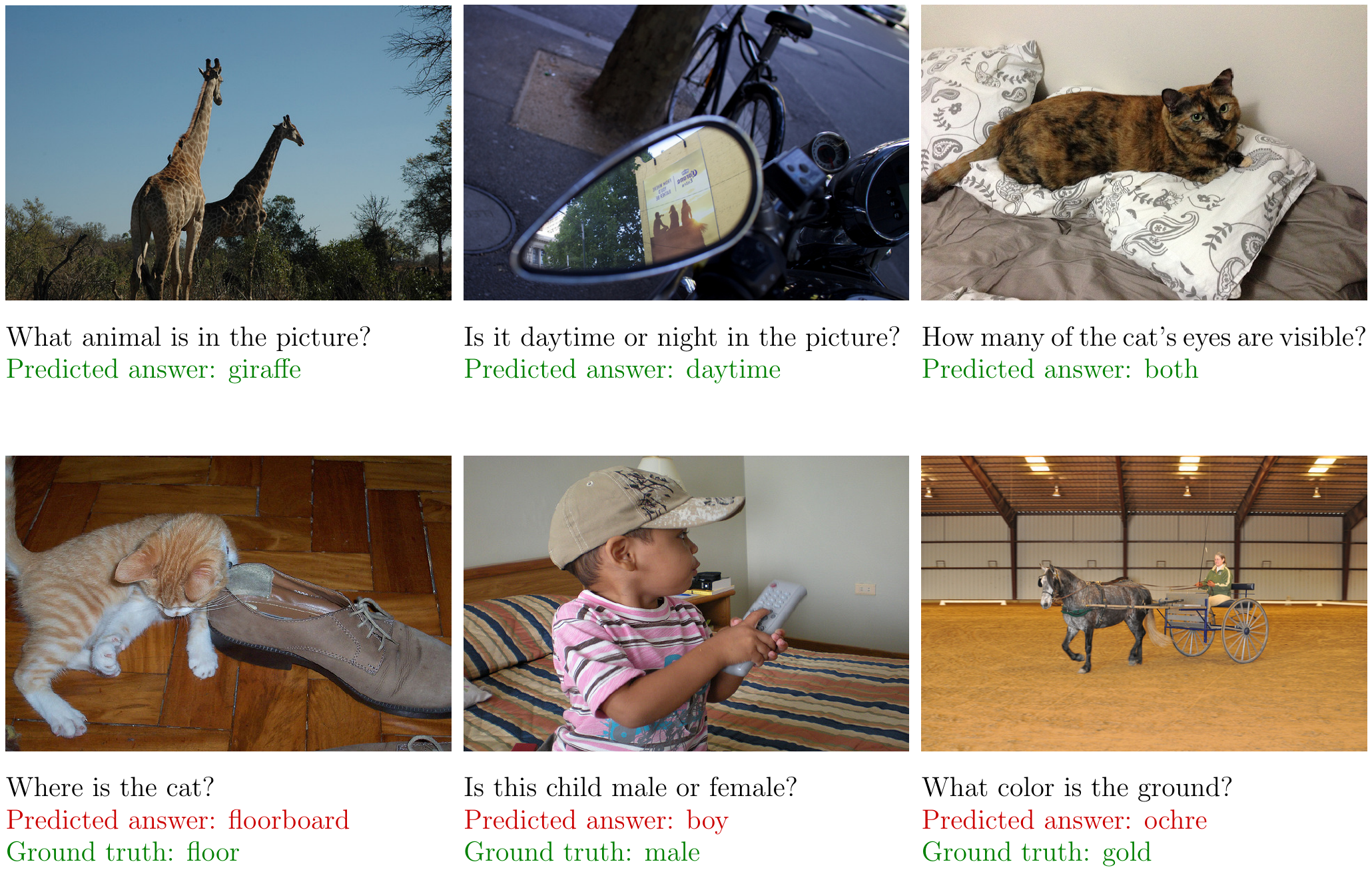}
  \caption{Examples of out-of-vocabulary predictions. The model works well when the ground truth answer has a clear and distinct pretrained embedding (top row), but fails to distinguish between synonymous answers (bottom row).}
  \label{fig:out-example}
\end{figure*}

We analyzed the predicted answers in out-of-vocabulary test setting to discover the cause of reduced performance and possible ways for improvement (\fig\ref{fig:out-example}). The reason for many failure cases is due to synonymous and/or related answers. When the representations of multiple candidate answers are close in the semantic space, it is difficult for the model to distinguish them, especially when they are both plausible for a given question.

Another important factor in the success of our method is how well the semantic space is covered by answers seen during training. For example, if the training questions all have similar answers, \eg different animal species, the model could generalize well to novel animals, but not as well to anything outside these. In otherwords, the model is perfectly capable of interpolation, but extrapolation remains a challenge.

The VQA v2 dataset was not originally designed to test the out-of-vocabulary prediction, and existing attempts to repurpose it (\eg \cite{teney2016zero}) all have notable issues. For our experiments, we created our own splits with novel answers in the test split, but we made no particular provision for an ``even'' coverage of semantic concepts with the training answers. These considerations suggest the need for a specific benchmark to allow a more rigorous evaluation of models designed for out-of-vocabulary and ``zero-shot'' VQA.

\end{document}